\documentclass[10pt,twocolumn,letterpaper]{article}

\usepackage[pagenumbers]{cvpr} 

\usepackage{graphicx}
\usepackage{amsmath}
\usepackage{amssymb}
\usepackage{booktabs}
\usepackage{tabulary}
\usepackage{multirow}
\usepackage[accsupp]{axessibility}

\usepackage[pagebackref,breaklinks,colorlinks]{hyperref}

\usepackage[capitalize]{cleveref}
\crefname{section}{Sec.}{Secs.}
\Crefname{section}{Section}{Sections}
\Crefname{table}{Table}{Tables}
\crefname{table}{Tab.}{Tabs.}

\newcolumntype{x}[1]{>{\centering\arraybackslash}p{#1pt}}
\newlength\savewidth
\newcommand{\tablestyle}[2]{\setlength{\tabcolsep}{#1}\renewcommand{\arraystretch}{#2}\centering\footnotesize}

\newcommand{\tocite}[1]{\textcolor{red}{[TOCITE]}}

\newcommand{\PAR}[1]{\vskip4pt \noindent{\bf #1~}}

\def\cvprPaperID{2925} 
\def\confName{CVPR}
\def\confYear{2023}

\begin{document}

\title{Representing Volumetric Videos as Dynamic MLP Maps}

\author{
Sida Peng$^*$
\quad
Yunzhi Yan$^*$
\quad
Qing Shuai
\quad
Hujun Bao
\quad
Xiaowei Zhou$^\dag$ \\[1.5mm]
State Key Lab of CAD\&CG, Zhejiang University
}
\maketitle

\begin{abstract}
    This paper introduces a novel representation of volumetric videos for real-time view synthesis of dynamic scenes.
    Recent advances in neural scene representations demonstrate their remarkable capability to model and render complex static scenes, but extending them to represent dynamic scenes is not straightforward due to their slow rendering speed or high storage cost.
    To solve this problem, our key idea is to represent the radiance field of each frame as a set of shallow MLP networks whose parameters are stored in 2D grids, called MLP maps, and dynamically predicted by a 2D CNN decoder shared by all frames.
    Representing 3D scenes with shallow MLPs significantly improves the rendering speed, while dynamically predicting MLP parameters with a shared 2D CNN instead of explicitly storing them leads to low storage cost.
    Experiments show that the proposed approach achieves state-of-the-art rendering quality on the NHR and ZJU-MoCap datasets, while being efficient for real-time rendering with a speed of 41.7 fps for $512 \times 512$ images on an RTX 3090 GPU.
    The code is available at \href{https://zju3dv.github.io/mlp_maps/}{https://zju3dv.github.io/mlp\_maps/}.
\end{abstract}

\let\thefootnote\relax\footnotetext{$^*$Equal contribution. $^\dag$Corresponding author.}
\section{Introduction}

Volumetric video captures a dynamic scene in 3D which allows users to watch from arbitrary viewpoints with immersive experience.
It is a cornerstone for the next generation media and has many important applications such as video conferencing, sport broadcasting, and remote learning.
The same as 2D video, volumetric video should be capable of high-quality and real-time rendering as well as being compressed for efficient storage and transmission.
Designing a proper representation for volumetric video to satisfy these requirements remains an open problem.

Traditional image-based rendering methods \cite{levoy1996light, davis2012unstructured, bansal20204d, yoon2020novel} build free-viewpoint video systems based on dense camera arrays.
They record dynamic scenes with many cameras and then synthesize novel views by interpolation from input nearby views. For these methods, the underlying scene representation is the original multi-view video.
While there have been many multi-view video coding techniques, the storage and transmission cost is still huge which cannot satisfy real-time video applications.
Another line of work \cite{collet2015high, dou2016fusion4d} utilizes RGB-D sensors to reconstruct textured meshes as the scene representation.
With mesh compression techniques, this representation can be very compact and enable streamable volumetric videos, but these methods can only capture humans and objects in constrained environments as reconstructing a high-quality renderable mesh for general dynamic scenes is still a very challenging problem. 

\begin{figure}[t]
\centering
\includegraphics[width=1\linewidth]{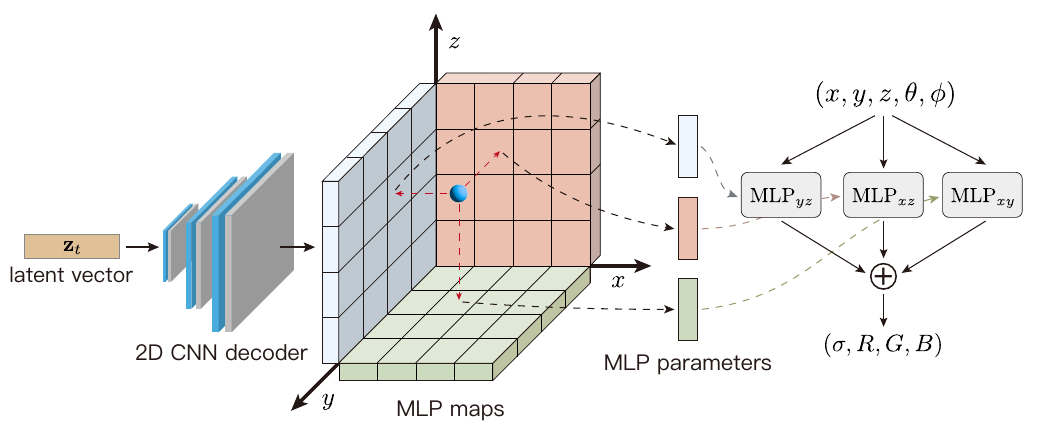}
\caption{\textbf{The basic idea of dynamic MLP maps.}
Instead of modeling the volumetric video with a big MLP network \cite{li2022neural}, we exploit a 2D convolutional neural network to dynamically generate 2D MLP maps at each video frame, where each pixel storing the parameter vector of a small MLP network.
This enables us to represent volumetric videos with a set of small MLP networks, thus significantly improving the rendering speed.
}
\label{fig:basic_idea}
\vspace{-4mm}
\end{figure}

Recent advances in neural scene representations \cite{lombardi2019neural, li2022neural, wang2022fourier} provide a promising solution for this problem.
They represent 3D scene with neural networks, which can be effectively learned from multi-view images through differentiable renderers.
For instance, Neural Volumes \cite{lombardi2019neural} represents volumetric videos with a set of RGB-density volumes predicted by 3D CNNs.
Since the volume prediction easily consumes large amount of GPU memory, it struggles to model high-resolution 3D scenes.
NeRF \cite{mildenhall2020nerf} instead represent 3D scenes with MLP networks regressing density and color for any 3D point, thereby enabling it to synthesize high-resolution images.
DyNeRF \cite{li2022neural} extends NeRF to model volumetric videos by introducing a temporal latent code as additional input of the MLP network.
A major issue of NeRF models is that their rendering is generally quite slow due to the costly network evaluation.
To increase the rendering speed, some methods \cite{garbin2021fastnerf, yu2021plenoctrees, wang2022fourier} utilize caching techniques to pre-compute a discrete radiance volume.
This strategy typically leads to high storage cost, which is acceptable for a static scene, but not scalable to render a volumetric video of dynamic scenes.

In this paper, we propose a novel representation of volumetric video, named dynamic MLP maps, for efficient view synthesis of dynamic scenes.
The basic idea is illustrated in Figure~\ref{fig:basic_idea}.
Instead of modeling a volumetric video with a single MLP network, we represent each video frame as a set of small MLP networks whose parameters are predicted by a per-scene trained 2D CNN decoder with a per-frame latent code.
Specifically, given a multi-view video, we choose a subset of views and feed them into a CNN encoder to obtain a latent code for each frame.
Then, a 2D CNN decoder regresses from the latent code to 2D maps, where each pixel in the maps stores a vector of MLP parameters. We call these 2D maps as MLP maps.
To model a 3D scene with the MLP maps, we project a query point in 3D space onto the MLP maps and use the corresponding MLP networks to infer its density and color values.

Representing 3D scenes with many small MLP networks decreases the cost of network evaluation and increases the rendering speed. This strategy has been proposed in previous works \cite{rebain2021derf, reiser2021kilonerf}, but their networks need to be stored for each static scene, which easily consumes a lot of storage to represent a dynamic scene.
In contrast to them, we use shared 2D CNN encoder and decoder to predict MLP parameters on the fly for each video frame, thereby effectively compressing the storage along the temporal domain.
Another advantage of the proposed representation is that MLP maps represent 3D scenes with 2D maps, enabling us to adopt 2D CNNs as the decoder instead of 3D CNNs in Neural Volumes \cite{lombardi2019neural}. This strategy leverages the fast inference speed of 2D CNNs and further decreases the memory requirement.

We evaluate our approach on the NHR and ZJU-MoCap datasets, which present dynamic scenes with complex motions.
Across all datasets, our approach exhibits state-of-the-art performance in terms of rendering quality and speed, while taking up low storage.
Experiments demonstrate that our approach is over 100 times faster than DyNeRF \cite{li2022neural}.

In summary, this work has the following contributions:

\begin{itemize}
\item A novel representation of volumetric video named dynamic MLP maps, which achieves compact representation and fast inference.
\item A new pipeline for real-time rendering of dynamic scenes based on dynamic MLP maps.
\item State-of-the-art performance in terms of the rendering quality, speed, and storage on the NHR and ZJU-MoCap datasets.
\end{itemize}

\begin{figure*}[t]
\centering
\includegraphics[width=1\textwidth]{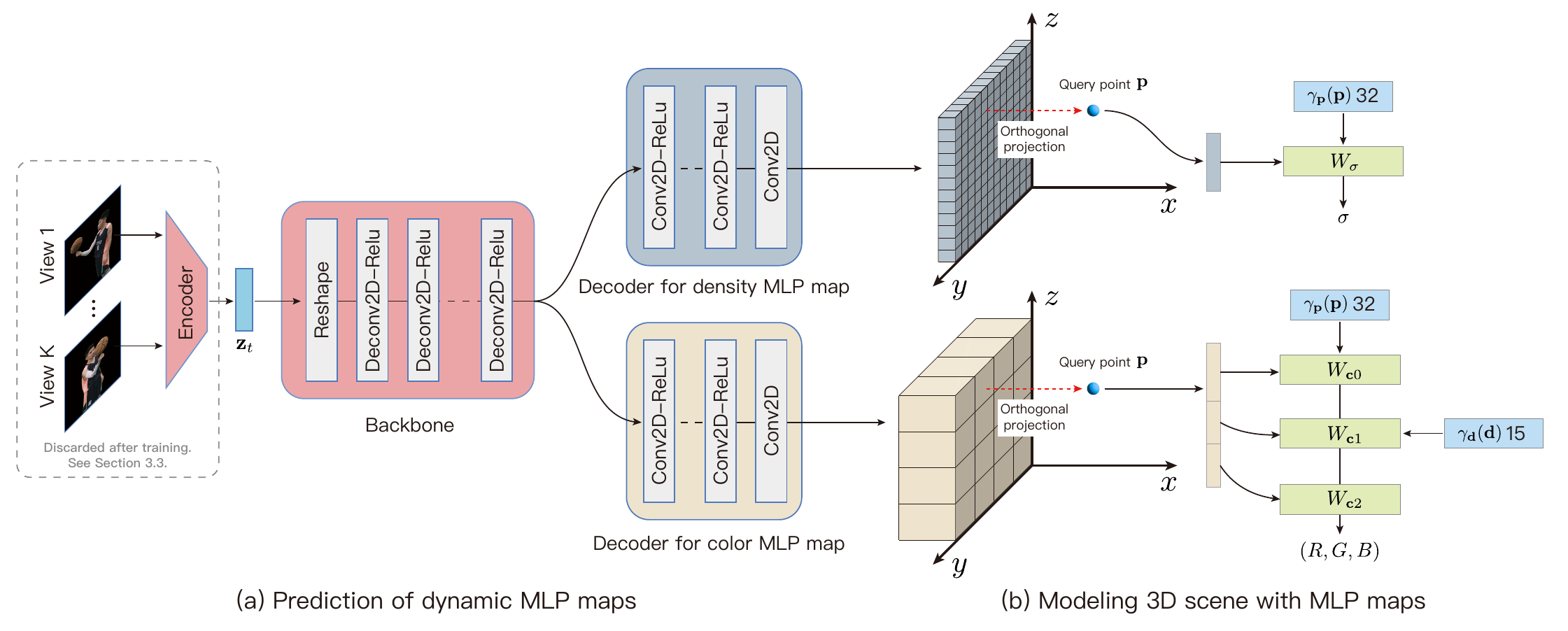}
\vspace{-2em}
\caption{\textbf{Illustration of dynamic MLP maps on the YZ plane.}
To model the volumetric video, a shared auto-encoder network predicts MLP maps to represent the 3D scene at each video frame.
(a) Specifically, for a particular video frame, an encoder network embeds a subset of input views into a latent vector, which is subsequentially processed by a decoder network to output density and color MLP maps.
(b) For any 3D query point, it is projected onto the 2D plane defined by the MLP map and retrieve the corresponding network parameters to construct the density and color head.
Finally, the resulted MLP network predicts the density and color for this point.
}
\label{fig:auto_encoder}
\vspace{-4mm}
\end{figure*}

\section{Related work}

\PAR{Traditional methods.}
Early works generally represent 3D scenes with multi-view images \cite{levoy1996light, davis2012unstructured} or explicit surface models \cite{collet2015high, dou2016fusion4d}.
Light field-based methods \cite{levoy1996light, davis2012unstructured, gortler1996lumigraph} build up a dense camera array to capture the target scene and synthesize novel views via light field interpolation techniques.
To reduce the number of input camera views, \cite{bansal20204d, yoon2020novel, lin2021efficient, johari2022geonerf, liu2022neural, wang2021ibrnet, suo2021neuralhumanfvv, chen2021mvsnerf} estimate the scene geometry from input views, then use the geometry to warp input views to the target view, and finally blend warped images to the target image.
These methods require storing captured RGB images for latter prediction, which could cause high storage costs.
Surface-based methods \cite{newcombe2015dynamicfusion, yu2018doublefusion, orts2016holoportation, collet2015high, dou2016fusion4d} leverage RGB-D sensors to reconstruct textured meshes to represent target scenes.
They utilize the depth sensors and multi-view stereo methods \cite{schonberger2016structure, zheng2014patchmatch} to obtain per-view depth images, and employ the depth fusion to obtain the scene geometry.
Although these methods are able to create high-quality streamable volumetric videos, the need of depth sensors limits them to work in constrained environments.

\PAR{Neural scene representations.}
These methods \cite{tucker2020single, wizadwongsa2021nex, mildenhall2020nerf} propose to represent 3D scenes with neural networks, which can be effectively learned from images through differentiable renderers.
Some surveys \cite{tewari2022advances, tewari2020state, xie2022neural} have summarized the neural scene representation methods.
NeRF \cite{mildenhall2020nerf, barron2021mip, barron2022mip} represents continuous volumetric scenes with MLP networks and achieves impressive rendering results.
To further improve the performance, some methods \cite{turki2022mega, tancik2022block, reiser2021kilonerf, rebain2021derf} decompose the scene into a set of spatial cells, each of which is represented by a NeRF network.
Block-NeRF \cite{tancik2022block} demonstrates that this strategy enables high-quality rendering for large-scale scenes.
KiloNeRF \cite{reiser2021kilonerf} represents each subregion with a smaller NeRF network and thus accelerates the rendering process.
Other methods speed up the rendering via caching techniques \cite{yu2021plenoctrees, hedman2021baking} or explicit representations \cite{liu2020neural, chen2022mobilenerf}.
Explicit representations are also used to increase the rendering quality, such as voxels \cite{liu2020neural, hao2021gancraft, xu20223d}, point clouds \cite{xu2022point}, and planes \cite{peng2020convolutional, chan2022efficient, chen2022tensorf, li2022fof}.
Previous methods \cite{chan2022efficient} have used tri-plane representations, where they leverage tri-plane feature maps to increase the model capacity.
Our model differs from them in that we predict tri-plane MLP maps instead of feature maps, and we can model the 3D scene with a single MLP map instead of three maps.

Recent methods \cite{lombardi2019neural, wang2021learning, lin2021deep, xing2022temporal} investigate the potential of neural 3D representations in representing volumetric videos.
To model high-resolution scenes, \cite{park2021nerfies, pumarola2021d, gao2021dynamic, du2021neural, peng2021animatable, liu2021neural, peng2021neural, xu2021h} extend NeRF to represent dynamic scenes.
\cite{li2021neural, xian2021space, li2022neural, gao2021dynamic, song2022nerfplayer} introduce temporal embedding as an additional input of NeRF network to describe dynamic scenes.
They typically adopt deep networks to achieve photorealistic rendering, resulting in slow rendering.
Another line of works \cite{pumarola2021d, park2021nerfies, park2021hypernerf, tretschk2021non, zhang2021editable} introduce a deformation field which maps points into a canonical space. However, as discussed in \cite{li2021neural, park2021nerfies}, deformation is difficult to optimize from images and thus struggle to represent scenes with large motions.

Some methods \cite{lombardi2019neural, wang2022fourier, garbin2021fastnerf, lombardi2021mixture, wang2022hvh} have explored the task of fast rendering of dynamic scenes.
Wang \etal \cite{wang2022fourier} utilize a generalized NeRF to build up a PlenOctree \cite{yu2021plenoctrees} for each video frame and compress these PlenOctrees into a Fourier PlenOctree with discrete Fourier transform.
To achieve high-quality compression, Fourier transform requires high dimensions of Fourier coefficients, which increases the storage cost.
As shown in \cite{wang2022fourier}, Fourier PlenOctree models a volumetric video of 60 frames with more than 7 GB storage cost.
In contrast, our approach represents dynamic scenes with 2D CNNs, resulting in a compact scene representation for real-time rendering.

\PAR{Hypernetworks.}
As discussed in Ha \etal \cite{ha2016hypernetworks}, a hypernetwork generates the parameters for another network.
There have been some works attempting to apply hypernetworks to computer vision tasks, such as semantic perception \cite{chen2020dynamic, tian2020conditional}, image generation \cite{karras2019style, karras2020analyzing}, view synthesis \cite{sitzmann2019scene, chen2022transformers}, human modeling \cite{wang2021metaavatar}, and inverse rendering \cite{maximov2019deep}.
In neural field-based methods \cite{sitzmann2019scene, wang2021metaavatar}, hypernetworks are generally utilized to generalize neural representations across a set of scenes.
To encode a set of scenes with a single network, SRN \cite{sitzmann2019scene} defines a set of latent vectors and uses an MLP network to map the latent vector to parameters of an MLP representing a specific scene.
Instead of regressing network weights with MLP networks, our approach proposes to use fully convolutional neural network to generates MLP maps, which efficiently produces a set of networks.

\section{Proposed approach}

Given a multi-view video captured by synchronized and calibrated cameras, our goal is to produce a volumetric video that takes up low disk storage and supports fast rendering.
In this paper, we propose a novel representation called dynamic MLP maps for volumetric videos and develop a pipeline for real-time view synthesis of dynamic scenes.
In this section, we first describe how to model 3D scenes with MLP maps (Section~\ref{sec:mlp_map}).
Then, Section~\ref{sec:vol_video} discusses how to represent volumetric videos with dynamic MLP maps.
Finally, we introduce some strategies to speed up the rendering process (Section~\ref{sec:speed}).

\subsection{Modeling 3D scenes with MLP maps}
\label{sec:mlp_map}

An MLP map is a 2D grid map with each pixel storing the parameters of an MLP network.
To represent 3D scene with MLP maps, we project any 3D point $\mathbf{p}$ to the 2D planes defined by MLP maps for querying the corresponding MLP parameters.
In practice, we align MLP maps with the axes of the coordinate frame, and point $\mathbf{p}$ is orthographically projected onto a canonical plane.
See Figure~\ref{fig:auto_encoder}(b) as an example of MLP maps on the YZ plane.
The projected query point is assigned to a parameter vector through spatial binning, which is dynamically loaded to the MLP network.
Here we adopt a small NeRF network to predict the density and color for the query point $\mathbf{p}$, which consists of one-layer density head and three-layer color head.
The networks encoded by the 2D maps together represent the whole scene.
Since each NeRF only describes a fraction of the target scene, our model can reach high-quality rendering with small MLPs.
In contrast to network sets used in previous methods \cite{turki2022mega, tancik2022block, reiser2021kilonerf, rebain2021derf}, the proposed MLP maps are in the format of 2D planes, enabling us to effectively and efficiently generate MLP parameters with 2D convolutional networks, which will be described latter.

When predicting the density and color for point $\mathbf{p}$, instead of directly passing the spatial coordinate into the network, we embed the input coordinate to a high-dimensional feature vector for better performance.
Specifically, we define three multi-level hash tables \cite{muller2022instant}: $\mathbf{h}_{xy}$, $\mathbf{h}_{xz}$, $\mathbf{h}_{yz}$.
Each hash table has a resolution of $L \times T \times F$, where $L$ is the number of hash table's levels, $T$ is the table size, and $F$ is the feature dimension.
To embed the input point $\mathbf{p}$, we project it onto three axis-aligned orthogonal planes, transform projected points to feature vectors using three multi-level hash tables, and aggregate the three feature vectors via summation.
The embedding process is formulated as:
\begin{equation}
    \resizebox{.9\hsize}{!}{
    $\gamma_{\mathbf{p}}^h(\mathbf{p}) = \eta(x, y, t; \mathbf{h}_{xy}) + \eta(x, z, t; \mathbf{h}_{xz}) + \eta(y, z, t; \mathbf{h}_{yz}),$
    }
\end{equation}
where $(x, y, z)$ is the coordinate of point $\mathbf{p}$, $t$ is the video frame, and $\eta$ is the encoding function, which obtains the feature vector from the hash table according to the input point (please refer to \cite{muller2022instant} for more details).
In addition, we follow EG3D \cite{chan2022efficient} to predict tri-plane feature maps with a 2D CNN and use them to assign a feature vector $\gamma_{\mathbf{p}}^t(\mathbf{p})$ to each 3D point.
We add up the hash table feature $\gamma_{\mathbf{p}}^h(\mathbf{p})$ and the tri-plane feature $\gamma_{\mathbf{p}}^t(\mathbf{p})$ to obtain the final feature vector $\gamma_{\mathbf{p}}(\mathbf{p})$.
The density $\sigma$ and color $\mathbf{c}$ is predicted via:
\begin{equation}
    (\sigma, \mathbf{c}) = M(\gamma_{\mathbf{p}}(\mathbf{p}), \gamma_{\mathbf{d}}(\mathbf{d})),
\end{equation}
where $M$ means the MLP network, and $\gamma_{\mathbf{d}}(\mathbf{d}))$ is the encoded viewing direction.
Figure~\ref{fig:auto_encoder}(b) visualizes the MLP network.
In practice, we implement $\gamma_{\mathbf{d}}$ as a positional encoding function \cite{mildenhall2020nerf}, and the 2D CNN shares parameters with the one used to dynamic MLP maps, whose detailed architecture is presented in Section~\ref{sec:implementation}.

\PAR{Orthogonal MLP maps.}
Experiments demonstrate that modeling scenes with MLP maps defined on one canonical plane struggles to give good rendering quality.
A reason is that scene content could be a high-frequency signal function along an axis, which makes the MLP map difficult to fit the scene content.
In such case, scene content may have lower frequency along another axes.
Inspired by \cite{peng2020convolutional, chan2022efficient, chen2022tensorf}, we define three MLP maps on three axis-aligned orthogonal planes.
For a query point, we first use the MLP maps to predict densities and colors $\{(\sigma_i, \mathbf{c}_i)\ | i = 1, 2, 3\}$ and then aggregate them via summation.
Figure~\ref{fig:basic_idea} illustrates the idea of summing the outputs of orthogonal signal functions.
The experimental results in Section~\ref{sec:ablation} show that three MLP maps defined on orthogonal planes perform better than three ones defined on the same planes.

\subsection{Volumetric videos as dynamic MLP maps}
\label{sec:vol_video}

Based on the MLP map, we are able to use a 2D CNN to represent the volumetric video.
Given a multi-view video, we leverage a 2D CNN to dynamically regress 2D maps containing a set of MLP parameters for each video frame, which model the geometry and appearance of 3D scene at corresponding time step.
As illustrated in Figure~\ref{fig:auto_encoder}(a), the network architecture is implemented as an encoder-decoder, where the encoder regresses latent codes from camera views and the decoder produces MLP maps based on latent codes.

Specifically, for a particular video frame, we select a subset of camera views and utilize 2D CNN encoder to convert them into a latent code following Neural Volumes \cite{lombardi2019neural}.
The latent code is designed to encode the state of the scene at the video frame and used for the prediction of MLP maps.
An alternative way to obtain latent codes is pre-defining learnable feature vectors for each video frame as in \cite{park2019deepsdf, martin2021nerf, bojanowski2017optimizing}.
The advantage of learning with an encoder network is that it implicitly shares the information across video frames, enabling joint reconstruction of video sequences rather than just per-frame learning.

Given the encoded latent vector, a 2D CNN decoder is employed to predict MLP maps.
Figure~\ref{fig:auto_encoder}(a) presents the schematic architecture of 2D CNN decoder.
Denote the latent vector as $\mathbf{z} \in \mathbb{R}^{D}$.
We first use a fully-connected network to map $\mathbf{z}$ to a 4096-dimensional feature vector and reshape the resulting vector as a $4 \times 4$ feature map with $256$ channels.
Then, a network with a series of deconvolutional layers upsamples it to a feature map of higher resolution $D \times D$. 
Based on the feature map, two subsequent convolutional networks are used to predict the density and color MLP maps, respectively.
The convolutional network consists of several convolutional layers.
By controlling the number and stride of convolutional layers, we control the resolution of the predicted MLP map.
Since the density MLP has fewer parameters than the color MLP, we can predict a higher resolution for the density MLP map, which leads to better performance, as demonstrated by experimental results in Section~\ref{sec:ablation}.

\vspace{1em}

\subsection{Accelerating the rendering}
\label{sec:speed}

Our approach represents 3D scenes with a set of small MLP networks.
Since these MLP networks are much smaller than that of DyNeRF \cite{li2022neural}, our network evaluation takes less time, enabling our approach to be much faster than DyNeRF.
To further improve the rendering speed, we introduce two additional strategies.

First, the encoder network can be discarded after training.
We use the trained encoder to compute the latent vector for each video frame, and store the resulted latent vectors instead of forwarding the encoder network every time, which save the inference time of the encoder.

Second, the number of network queries is reduced by skipping the empty space.
To this end, we compute a low-resolution 3D occupancy volume per video frame, where each volume voxel stores a binary value indicating whether the voxel is occupied.
The occupancy volume is extracted from the learned scene representation.
We mark a voxel as occupied when its evaluated density is higher than a threshold.
Since the occupancy volume has a low resolution and is stored in the binary format, the occupancy volumes of a 300-frame video only take up about 8 MB storage.
During inference, the network evaluation is only performed on occupied regions indicated by the occupancy volume.
In practice, we will run the density evaluation before the color evaluation to further reduce the number of color evaluations.
Specifically, we first evaluate the densities of sampled points within occupied voxels, which are used to calculate the composition weights as defined in the volume rendering \cite{mildenhall2020nerf,li2022nerfacc}.
If the weight of a point is higher than a threshold, we then predict its color value, otherwise we skip this point during the volume rendering.

The complete rendering pipeline and the corresponding implementation details are presented in the supplementary material.
The code will be released for the reproducibility.
\section{Implementation details}\label{sec:implementation}

\PAR{Network architecture.}
Our approach adopts an encoder with seven convolutional layers, which takes three $512 \times 512$ images as input and converts them into a 256-dimensional latent vector following \cite{lombardi2019neural}.
The decoder network consists of a backbone network and two prediction heads.
The backbone network has six deconvolutional layers and upsamples the input $4 \times 4$ feature map to a $256 \times 256$ backbone feature map.
We use a convolutional layer to regress from the backbone feature map to a 96-channel feature map and reshape it to three 32-channel planes.
The prediction head for density MLP is implemented as a convolutional layer with a stride of 1 and outputs a $256 \times 256$ MLP map, each having 32 parameters ($32 \times 1$).
The prediction head for color MLP applies four convolutional layers with a stride of 2 and one convolutional layer with a stride of 1 to the input feature map, resulting in a $16 \times 16$ MLP map, each having 2624 parameters ($32 \times 32 + (32 + 15) \times 32 + 32 \times 3$) with ReLU internal activation.
For each multi-level hash table, we follow \cite{muller2022instant} to set $L = 19$, $T = 16$ and $F = 2$ .

\PAR{Storage cost.}
For an input video of 300 frames, the storage cost of latent vectors, MLP maps decoders, multi-level hash tables and occupancy volumes in our model are 300 KB, 103MB, 131MB and 8MB, respectively. Note that, the encoder network are not stored thanks to our accelerating strategies in Section \ref{sec:speed}.

\PAR{Training.}
The networks are optimized to minimize the MSE loss that calculates the difference between rendered and observed images. 
The Kullback-Leibler divergence loss in \cite{lombardi2019neural} is also used for training.
More details of loss functions can be found in the supplementary material.
During optimization, we set the batch size as eight and train the model on one RTX 3090 GPU, which takes about 16 hours.
The learning rate is set as $5e^{-4}$ and $5e^{-3}$ for auto-encoder network and hashtable respectively and decays exponentially by 0.1 every 400 epochs.

\vspace{1em}
\section{Experiments}

\subsection{Datasets}

To evaluate the performance of our approach, we conduct experiments of the ZJU-MoCap \cite{peng2021neural} and NHR \cite{wu2020multi} datasets.
Both two datasets capture foreground dynamic scenes within bounded regions using multiple synchronized cameras, and the recorded scenes exhibit large motions.
On ZJU-MoCap dataset, we uniformly select 11 cameras for training and use the remaining views for evaluation. All video sequences have a length of 300 frames.
On NHR dataset, 90 percent of cameras are taken as training views and the other views are kept for evaluation. We select 100 frames from captured videos to reconstructing volumetric videos for each scene.
The two datasets provide segmentation masks for foreground dynamic scenes.
Based on foreground masks, we produce visual hulls of foreground scenes and calculate axis-aligned bounding boxes enclosing scenes, which are used in our approach.

\subsection{Ablation studies}\label{sec:ablation}

\begin{table}
\begin{center}
\scalebox{0.64}{
\begin{tabular}{x{42}|x{68}x{64}x{68}x{68}}
\toprule
Methods & Baseline1 & C-NeRF & Ours-Single & Ours \\[.1em]
\midrule
\multirow{3}{*}{Description} & \multirow{3}{7em}{feature tri-planes + single NeRF}  & \multirow{3}{7em}{feature volumes + single NeRF} & feature tri-planes + hash tables + single NeRF & feature tri-planes + hash tables + MLP maps \\
\midrule
LPIPS & 0.072 & 0.076 & 0.065 & \textbf{0.058} \\
\bottomrule
\end{tabular}
} 
\vspace{-1.em}
\end{center}
\caption{\textbf{Analyzing what contributes to the rendering quality}. Metrics are averaged over two scenes from ZJU-MoCap and NHR.}
\vspace{-2mm}
\label{table:comparison_baselines}
\end{table}

\begin{table}
\begin{center}
\scalebox{0.83}{
\tablestyle{4pt}{1.05}
\begin{tabular}{l|x{42}|c|c}
\toprule
& LPIPS $\downarrow$
& \begin{tabular}{c} Rendering time \\ (milliseconds) \end{tabular}  $\downarrow$
& \begin{tabular}{c} Storage \\ (MB) \end{tabular}  $\downarrow$
\\[.1em]
\midrule
(1) Color map size $1 \times 1$ & 0.069 & 22 & 489 \\[.1em]
(2) Color map size $4 \times 4$ & 0.065 & 23 & 324 \\[.1em]
(3) Color map size $32 \times 32$ & 0.060 & 27 & 208 \\[.1em]
(4) Density map Size $16 \times 16$ & 0.064 & 25 & 250 \\
\midrule
(5) Single shared MLP & 0.065 & 90 & 133 \\[.1em]
(6) Single shared MLP w/o ESS & 0.065 & 1345 & \textbf{125} \\[.1em]
\midrule
(7) XY MLP map & 0.072 & \textbf{19} & 206 \\[.1em]
(8) w/o orthogonal signal & 0.067 & 24 & 242\\[.1em]
\midrule
(9) w/o ESS & \textbf{0.058} & 144 & 237 \\[.1em]
\midrule
(10) Complete model & \textbf{0.058} & 24 & 242 \\
\bottomrule
\end{tabular}
}
\caption{\textbf{Ablation studies of our model.} Metrics are averaged over two scenes from the ZJU-MoCap and NHR datasets. We test the time of rendering a $512 \times 512$ image on one RTX 3090 GPU. See Section~\ref{sec:ablation} for detailed descriptions.}
\vspace{-3mm}
\label{table:ablation}
\end{center}
\end{table}

We validate our algorithm's design choices on two scenes from the ZJU-MoCap and NHR datasets.
Tables \ref{table:comparison_baselines} and \ref{table:ablation} present the quantitative results.

\PAR{What contributes to our rendering quality.}
Table~\ref{table:comparison_baselines} lists the results of ablation studies.
``Baseline1'' consists of tri-plane feature maps and a single shared MLP for all video frames.
The MLP has the same network as the original NeRF.
``Ours-Single'' is the model in Table~\ref{table:ablation} (5).
Table~\ref{table:comparison_baselines} indicates two components important to our rendering quality: 1) Hash tables (Ours-Single vs. C-NeRF \cite{wang2021learning}). 2) MLP maps (Ours vs. Ours-Single). Using MLP maps allows each MLP to represent a small region, in contrast to the single NeRF modeling the whole volumetric video.

\PAR{The resolution of MLP maps.}
Experimental results in rows 1, 2, and 4 of Table~\ref{table:ablation} show that decreasing the resolution undermines the rendering performance.
Note that, the storage cost increases when the resolution decreases. 
This is because that we add more convolutional layers for the downsampling, as described in Section~\ref{sec:implementation}.

\begin{figure}[t]
\centering
\includegraphics[width=1\linewidth]{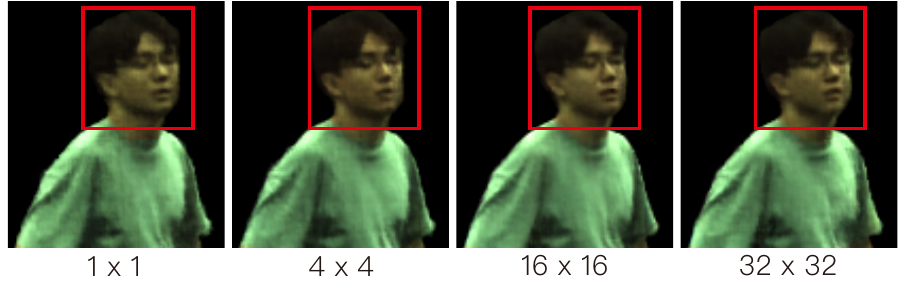}
\vspace{-1.5em}
\caption{\textbf{Qualitative results of models with different color MLP map resolutions} on the ZJU-MoCap dataset.}
\label{fig:ablation_resolution}
\vspace{-1.5em}
\end{figure}

Results of row 3 show that increasing the resolution does not necessarily improves the rendering quality.
A plausible explanation is that the increased resolution make the model more difficult to train for two factors.
First, the higher resolution of MLP map means more parameters to be optimized.
Second, increasing the resolution need to decrease the number of query points fed into each color MLP during each training iteration due to memory limit, thereby decreasing the batch size.
Figure~\ref{fig:ablation_resolution} shows some visual results.

It is interesting to analyze why the high-resolution density map is trainable while the color map is not.
The reason is that density MLP has 32 parameters, and color MLP has 2624 parameters, which means that $256 \times 256$ density MLP map has similar number of parameters to $28 \times 28$ color MLP map.
In addition, the scene geometry generally has lower frequency than the appearance.
Therefore, the high-resolution density MLP maps are easier to train.

The results in rows 1-4 of Table \ref{table:ablation} also indicate that the resolution of MLP maps does not affect the rendering speed much.
This can be attributed to that the number of MLP evaluations is not related to the map resolution.

\begin{figure}[t]
\centering
\includegraphics[width=1\linewidth]{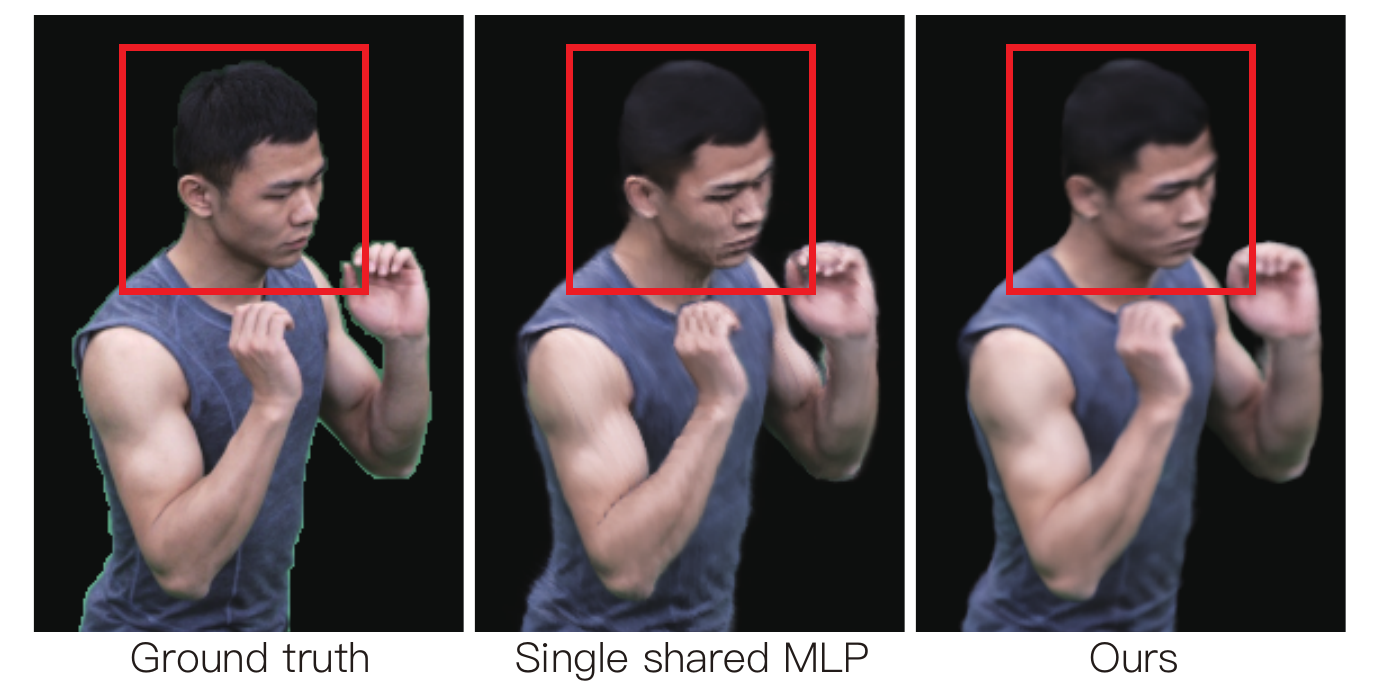}
\caption{\textbf{Comparisons between dynamic MLP maps and a single shared MLP} on the NHR dataset.}
\label{fig:ablation_single_map}
\end{figure}

\begin{figure}[t]
\centering
\includegraphics[width=1\linewidth]{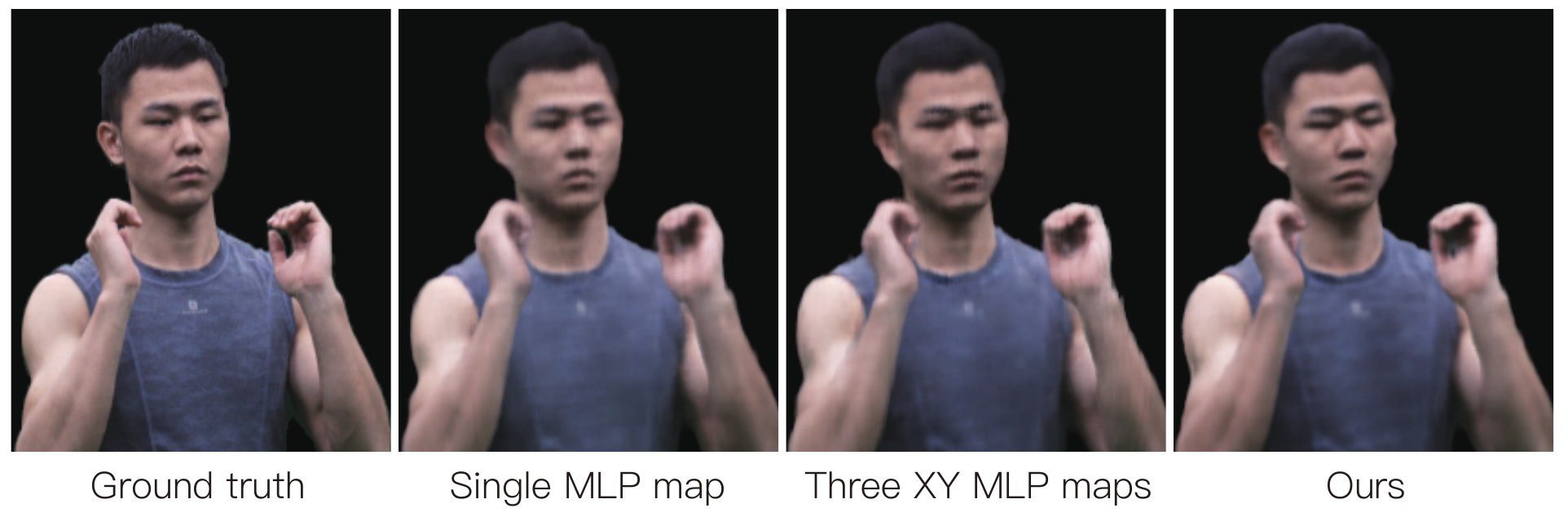}
\vspace{-1.5em}
\caption{\textbf{Ablation studies on the orthogonal MLP maps.} The results indicate that orthogonal MLP maps improves the quality.}
\label{fig:ablation_orthogonal}
\end{figure}

\PAR{Our model with a shared MLP.}
Rows 5-6 of Table~\ref{table:ablation} means that we replace the dynamic MLP maps with a single shared MLP for all video frames.
This MLP network is implemented as the original NeRF network \cite{mildenhall2020nerf} except that the input is the high-dimensional feature vector $\gamma_{\mathbf{p}}(\mathbf{p})$ in Section~\ref{sec:vol_video}.
Our complete model has better rendering quality and faster rendering speed than the model with a single shared MLP network.
The results of row 6 indicate that, with the empty space skipping, the rendering speed of model ``Single shared MLP'' increases much.
But it is still much slower than our complete model.

\PAR{The influence of using orthogonal MLP maps.}
Row ``XY MLP map'' in Table~\ref{table:ablation} means that we represent the scene with a single dynamic MLP map defined on the XY plane.
The results show the rendering quality degrades much.
Row ``w/o orthogonal signal'' indicates that we adopts three dynamic MLP maps, which are all defined on the XY plane.
The resulting model also shows a bad rendering quality.

\PAR{Analysis of our rendering time.}
Row ``w/o ESS'' in Table~\ref{table:ablation} investigates the effect of the empty space skipping.
The results show that this technique improves the rendering speed 6 times while only adding 5 MB on the storage cost.
Here we analyze the running time of each model component.
The inference time of MLP maps decoder and querying multi-level hash tables are about 4ms and 7ms, respectively. Our color and density MLP maps take only 12ms in total, in comparison with over 80ms taken by a single shared MLP with empty space skipping.

\begin{figure*}[t]
\centering
\includegraphics[width=1\textwidth]{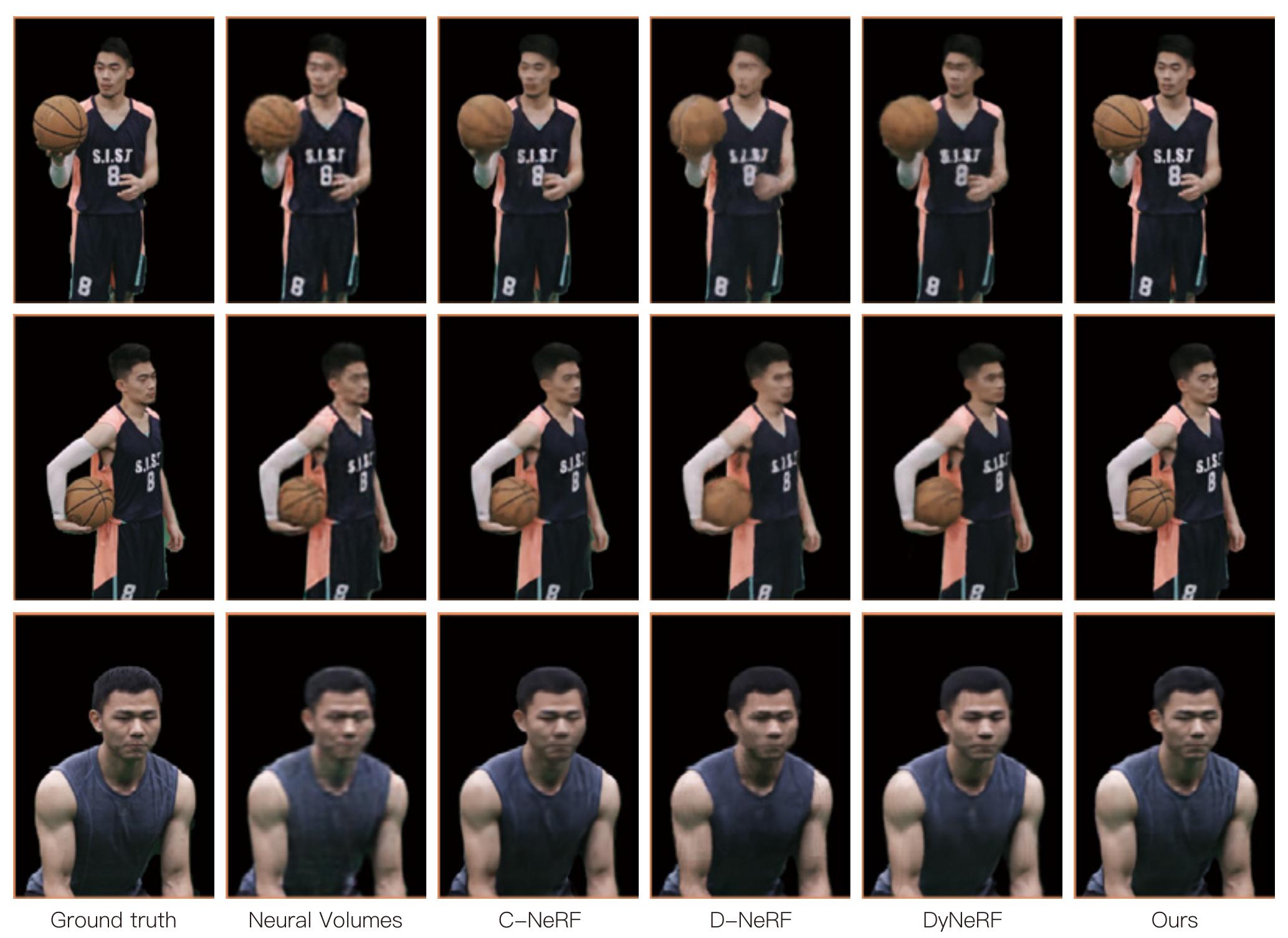}
\caption{\textbf{Qualitative comparisons on the NHR dataset.} Our approach significantly outperforms other methods. The results in the first rows indicates that we can synthesize detailed texture of basketballs, while the rendering results of other methods are blurry.}
\label{fig:nhr_results}
\end{figure*}

\subsection{Comparisons with the state-of-the-art methods}

\begin{figure*}[t]
\centering
\includegraphics[width=1\textwidth]{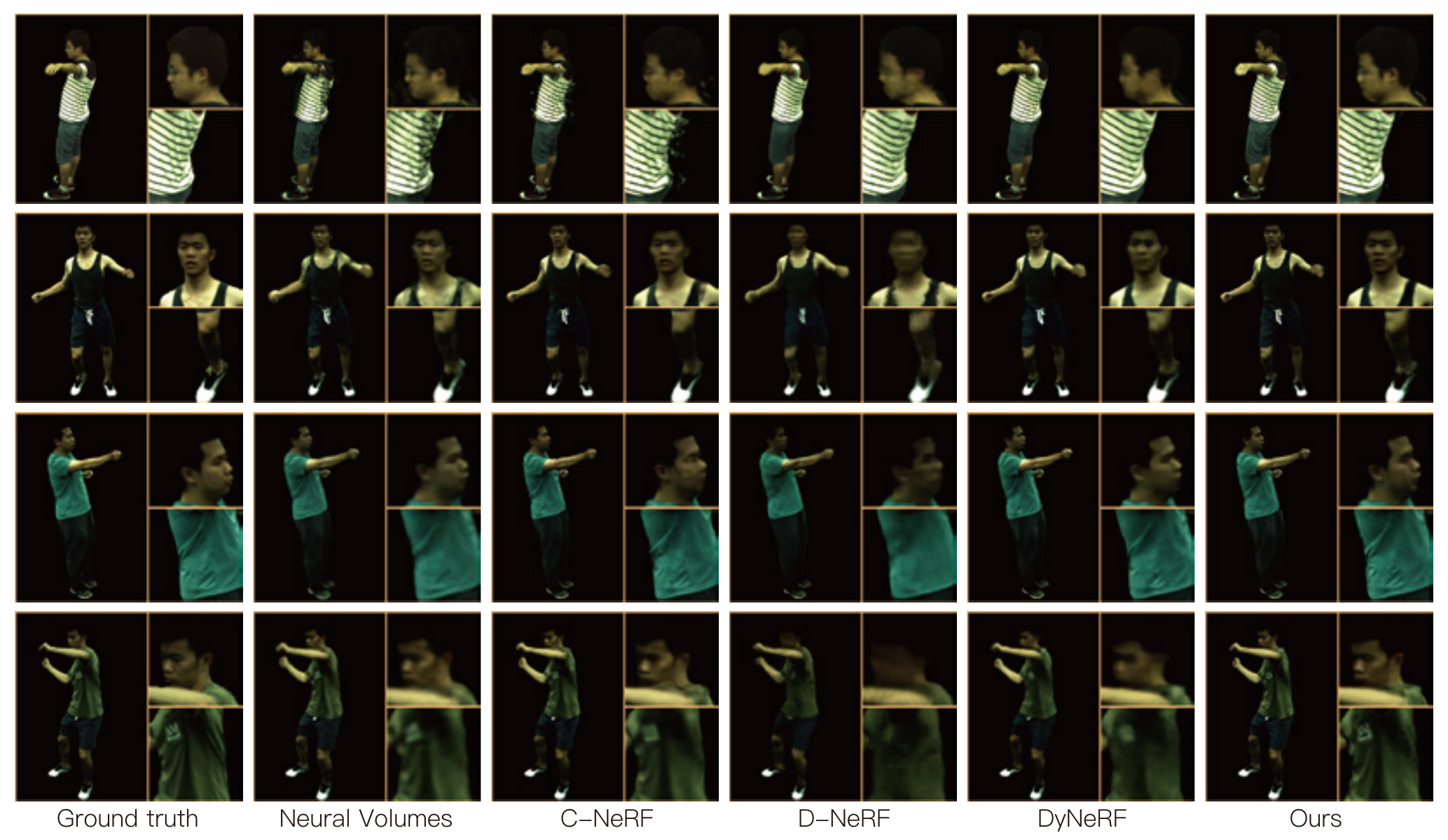}
\caption{\textbf{Qualitative comparisons in the ZJU-MoCap dataset.} Our proposed model can generate sharp details, as shown by the results of last two rows. In contrast, other methods tend to render smooth images.}
\label{fig:comparison_results}
\vspace{4mm}
\end{figure*}

\paragraph{Baseline methods.}
We compare our approach with four recent methods.
(1) Neural Volumes (NV) \cite{lombardi2019neural} uses a 3D deconvolutional network to produce RGB-Alpha volumes as the volumetric video.
(2) C-NeRF \cite{wang2021learning} is a follow-up work of Neural Volumes, which adopts a 3D CNN to produce feature volumes and appends a NeRF-like MLP to predict radiance fields.
(3) D-NeRF \cite{pumarola2021d} decomposes the dynamic scene into a static scene in the canonical space and a deformation field.
(4) DyNeRF \cite{li2022neural} represents the volumetric video with a NeRF-like MLP, which takes spatial coordinate and temporal embedding as input and predict the radiance field at a particular video frame.
Since C-NeRF and DyNeRF do not release the code, we re-implement them for the comparison.
We do not compare with Neural Body \cite{peng2021neural} and MVP \cite{lombardi2021mixture}, because their algorithms take tracked meshes as input, making the comparison unfair.

\begin{table}
\begin{center}
\scalebox{0.85}{
\tablestyle{4pt}{1.05}
\begin{tabular}{l|c|c|c|c|c}
\toprule
& PSNR $\uparrow$
& SSIM $\uparrow$
& LPIPS $\downarrow$
& \begin{tabular}{c} Rendering time \\ (milliseconds) \end{tabular} $\downarrow$
& \begin{tabular}{c} Storage \\ (MB) \end{tabular} $\downarrow$
\\[.1em]
\midrule
NV \cite{lombardi2019neural} & 30.86 & 0.941 & 0.130 & 73 & 658  \\[.1em]
C-NeRF \cite{wang2021learning} & 31.32 & 0.949 & 0.102 & 1969 & 1019 \\[.1em]
D-NeRF \cite{pumarola2021d} & 29.25 & 0.920 & 0.150 & 2303 & \textbf{4}\\[.1em]
DyNeRF \cite{li2022neural} & 30.87  & 0.943 & 0.118 & 5195 & 12 \\[.1em]
Ours & \textbf{32.20} & \textbf{0.953} & \textbf{0.080} & \textbf{33} & 239\\[.1em]
\bottomrule
\end{tabular}
}
\caption{\textbf{Quantitative results on the NHR dataset.} Metrics are averaged over all scenes. This dataset includes $512 \times 612$ images and $384 \times 512$ images.}
\vspace{-5mm}
\label{table:nhr}
\end{center}
\end{table}

\begin{table}
\begin{center}
\scalebox{0.85}{
\tablestyle{4pt}{1.05}
\begin{tabular}{l|c|c|c|c|c}
\toprule
& PSNR $\uparrow$
& SSIM $\uparrow$
& LPIPS $\downarrow$
& \begin{tabular}{c} Rendering time \\ (milliseconds) \end{tabular} $\downarrow$
& \begin{tabular}{c} Storage \\ (MB) \end{tabular} $\downarrow$
\\[.1em]
\midrule
NV \cite{lombardi2019neural} & 28.12 & 0.934 & 0.131 & 49 & 658\\[.1em]
C-NeRF \cite{wang2021learning} & 29.79 & 0.959 & 0.077 & 1313 & 1019\\[.1em]
D-NeRF \cite{pumarola2021d} & 27.08 & 0.922 & 0.139 & 1534 & \textbf{4}\\[.1em]
DyNeRF \cite{li2022neural} & 29.88 & 0.959 & 0.087 & 3452 & 12\\[.1em]
Ours & \textbf{30.17} & \textbf{0.963} & \textbf{0.068} & \textbf{24} & 245 \\[.1em]
\bottomrule
\end{tabular}
}
\caption{\textbf{Comparison on the ZJU-MoCap dataset.} Metrics are averaged over all scenes. The image resolution is $512 \times 512$.}
\vspace{-2em}
\label{table:zju_mocap}
\end{center}
\end{table}

Tables~\ref{table:nhr}, \ref{table:zju_mocap} list the comparison of our method with \cite{lombardi2019neural, wang2021learning, pumarola2021d, li2022neural} in terms of rendering quality, rendering speed and storage. We adopt PSNR, SSIM and LPIPS\cite{zhang2018unreasonable} as metrics to evaluate rendering quality. For all metrics, our method achieves the best performance among all the methods. Moreover, our model renders two magnitude faster than C-NeRF, D-NeRF and DyNeRF while maintaining reasonable storage cost. Although our approach is only twice as fast as Neural Volumes due to its explicit scene representation, our rendering accuracy is significantly better.
Figures~\ref{fig:nhr_results}, \ref{fig:comparison_results} show the qualitative results of our method and baselines. Limited by low volume resolution, Neural Volumes gives blurry results. D-NeRF has similar problems as deformation fields are difficult to learn when complex scene motions exist. C-NeRF and DyNeRF tend to lose details in high frequency regions of the image because of their single shared MLP model structures. In contrast, our method generates photorealistic novel view results.

\section{Conclusion}

We proposed a novel neural representation named dynamic MLP maps for volumetric videos.
The key idea is utilizing a shared 2D CNN to predict 2D MLP maps of each video frame, which store the parameters of MLP networks at pixels.
To model 3D scene with MLP maps, our approach regresses the density and color of any 3D point by projecting it onto the 2D planes defined by MLP maps and queries the corresponding the MLP networks for the prediction.
Experiments demonstrated that our approach achieves competitive rendering quality and low storage costs, while being efficient for real-time rendering.

\PAR{Limitations.}
1) Common videos are more than a few minutes.
However, this work only deals with videos of 100 to 300 frames, which are relatively short, thus limiting the applications.
How to model a long volumetric video remains an interesting problem.
2) Since we use multi-level hash tables, the storage cost of our model will increase as the video length increases.
3) The proposed representaion requires dense camera views for training, similar to \cite{li2022neural, lombardi2019neural, wang2021learning}.
Recovering free-viewpoint videos from sparse-view videos is also an open problem.

\vspace{1em}
\noindent\textbf{Acknowledgements.} 
This work was supported by the Key Research Project of Zhejiang Lab (No. K2022PG1BB01), NSFC (No. 62172364), and Information Technology Center and State Key Lab of CAD\&CG, Zhejiang University.

{\small
\bibliographystyle{ieee_fullname}
\bibliography{egbib}
}

\newpage
\setcounter{section}{0}
\twocolumn[{
 \centering
 \LARGE Supplementary Material for\\ Representing Volumetric Videos as Dynamic MLP Maps \\[1.5em]
}]

\section{More Implementation details}

\PAR{Empty space skipping.}
Resolution of the occupancy volume per video frame is set to $24 \times 24 \times 48$. 
During evaluation, we divide each volume cell into a subgrid of $5 \times 5 \times 5$ and calculate the density of each subgrid point following \cite{reiser2021kilonerf}. A volume cell is considered as occupied if there exists a subgrid point inside with density above threshold $\tau_1$ ($\tau_1 = 5$ in all our experiments). 

\PAR{Rendering pipeline.}
During training, we uniformly sample 64 query locations along each camera ray within the scene bound. We do not use hierarchical sampling in \cite{mildenhall2020nerf}.
For each query point, we first project it onto three axis-aligned orthogonal MLP maps which is predicted by the 2D CNN decoder to get the corresponding MLP parameters.
Then, we embed the query point to high-dimensional feature vector using the multi-level hash tables and feed it into the MLP network to predict the color and density. 
To efficiently evaluate multiple MLP networks, we develop a custom PyTorch layer \cite{Paszke_PyTorch_An_Imperative_2019} based on the library MAGMA \cite{abdelfattah2017novel}, following KiloNeRF \cite{reiser2021kilonerf}.
We obtain the final result by combining the prediction from each MLP map via elementwise summation.
Pixel color is rendered using the differentiable volume rendering following \cite{mildenhall2020nerf}, which is defined as:
\begin{equation}
\label{volume rendering}
\begin{aligned}
\tilde{C} (\mathbf{r}) =& \sum_{i = 1}^{M} \omega_i \mathbf{c}_i,  
\\
\omega_i =& ~T_{i}(1 - \text{exp}(-\sigma_{i}\delta_{i})),
\\
T_{i} =& ~\text{exp}(-\sum_{j=1}^{i-1} \sigma_{j}\delta_{j})
\end{aligned}
\end{equation}
where $c_{i}, \sigma_{i}$ denotes color and density predicted by MLP maps, $T_{i}$ is the accumulated transmittance, $\omega_i$ is the composition weight, and $\delta_{i}$ denotes the ray step size.

During inference, we set the sampling step size to be $\frac{1}{256}^{th}$ the diagonal distance of the scene bound. Sampling step size denotes distance between adjacent sampling locations on a ray. For each ray, we sample points only in occupied voxel with the pre-computed occupancy volume.
To further speed up the rendering process, we first obtain the density values and compute the accumulated transmittance and composition weight.
Sampling points are filtered for the color evaluations with threshold $\tau_2$ on composition weight ($\tau_2 = 1e^{-3}$ in all our experiments), which means we ignore a query point if its weight is below $\tau_2$. Colors and densities are predicted in the same way as in the training stage, and the final predicted color is computed via volume rendering as Eq.~\eqref{volume rendering}.
The rendering pipeline is implemented based on \cite{li2022nerfacc}.

\PAR{Loss function.} The loss function is defined as:
\begin{equation}
    L = L_c + \lambda_{KL}L_{KL}.
\end{equation}
Here $L_c$ is the error between rendered pixel color $\tilde{C}(\mathbf{r})$ and observed pixel color $C(\mathbf{r})$:
\begin{equation}
    L_c = \sum_{\mathbf{r} \in \mathcal{R}} || \tilde{C}(\mathbf{r}) - C(\mathbf{r}) ||^2_2,
\end{equation}
where $\mathcal{R}$ means the set of camera rays. The Kullback-Leibler divergence loss $L_{KL}$ follows in the formation in \cite{lombardi2019neural}. We set $\lambda_{KL} = 1e^{-6}$ in all our experiments.
If the target scenes contain only foreground objects, we also add the mask loss to help the training, which measures the error between ground truth foreground mask $\tilde{M}(\mathbf{r})$ and rendered image opacities $M(\mathbf{r})$:
\begin{equation}
    L_m = \sum_{\mathbf{r} \in \mathcal{R}} || \tilde{M}(\mathbf{r}) - M(\mathbf{r}) ||^2_2.
\end{equation}
The weight of the mask loss is set as 0.1 in experiments.

\section{Results on the data of Neural Volumes}

Neural Volumes \cite{lombardi2019neural} has released a video sequence in their paper, which records the floating dry ice.
We found that there is a moving human in the background of some camera views, which is different from the data presented in the paper of Neural Volumes.
Since Neural Volumes does not describe the training frames and camera views, we selected a 100-frame video clip, ranging from frame 16120 to 16417. We test our model on camera 400015 and train on the remaining cameras except 400055 and 400070.

\begin{figure}
  \centering
  \includegraphics[width=\linewidth]{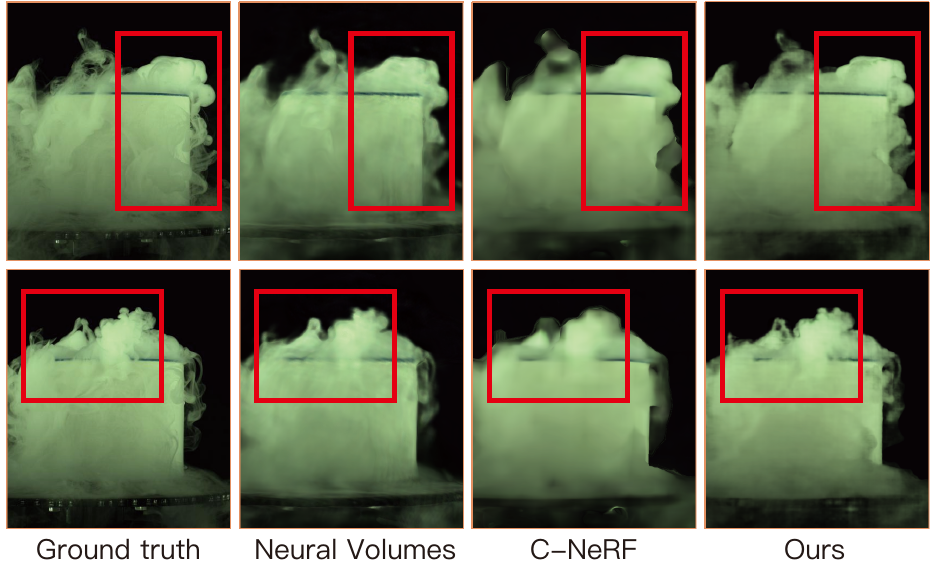}
  \caption{\textbf{Qualitative results on the data of Neural Volumes.} We render more photorealistic images than baseline methods.}
  \label{fig:nv}
  \end{figure}

The results in Table~\ref{table:nv} demonstrate that our method outperforms Neural Volumes and C-NeRF \cite{wang2021learning}.
Figure~\ref{fig:nv} presents the qualitative comparisons.

\begin{table}
\begin{center}
\scalebox{1}{
\tablestyle{4pt}{1.05}
\begin{tabular}{l|c|c|c}
\toprule
& PSNR $\uparrow$
& SSIM $\uparrow$
& LPIPS $\downarrow$
\\[.1em]
\midrule
NV \cite{lombardi2019neural} & 32.01 & 0.951 & 0.218 \\[.1em]
C-NeRF \cite{wang2021learning} & 31.84 & \textbf{0.953} & 0.227 \\[.1em]
Ours & \textbf{32.25} & 0.948 & \textbf{0.198} \\
\bottomrule
\end{tabular}
}
\caption{\textbf{Comparison on the data of Neural Volumes.} Our method outperforms Neural Volumes and C-NeRF quantitatively.}
\label{table:nv}
\end{center}
\end{table}

\section{Discussion}

\PAR{Dynamic MLP maps vs. per-frame KiloNeRF.}
An alternative way to represent real-time volumetric video is storing a sequence of per-frame KiloNeRF \cite{reiser2021kilonerf}.
This scheme makes the storage and training time increase linearly with the number of video frames.
For example, given a 300-frame video, we need to store 300 KiloNeRF models.
Consider that a KiloNeRF model requires 20 training hours and 30 MB.
300 KiloNeRF models would take 6000 hours for training and consume 9 GB in storage.
In contrast, our method requires 16 hours for training and takes up about 240 MB in storage.

\PAR{Motivation of using two parameter sets.}
We represent the volumetric video with two set of parameters: (1) hash tables and (2) a CNN that generates the MLP maps.
The motivation of using a hybrid of hash tables and MLP maps to achieve both high quality and efficiency.
1) Why MLP maps: Only using hash tables needs a relatively large MLP to achieve high quality, which will be slow due to the costly network evaluation, as suggested by the rendering speed of DyNeRF and C-NeRF.
Using MLP maps increases the rendering speed with small MLPs.
2) Why hash tables: compared to MLP maps alone, hash tables improve the rendering quality, as shown in the section of ablation studies.
We will make the motivation clearer in the revised paper.

\PAR{Why use MLP maps instead of feature maps.}
We use MLP maps to improve rendering speed.
Table 1 in the section of ablation studies indicates that using feature maps instead of MLP maps needs a large MLP to achieve high quality, which will be slow.
Feature maps can work together with MLP maps, but it does not perform better than hash tables, as demonstrated by the results of ablation studies.

\PAR{Contribution to rendering speed.}
When the ESS is not used, the rows 6 and 9 in Table 1 of ablation studies shows that MLP maps make the rendering 9x faster.
When the ESS is used, the results of ablation studies indicate that MLP maps make the rendering 4x faster.
ESS is a super effective technique. We believe that achieving another 4x speedup on top of it by a novel representation design is non-straightforward, beneficial to the community, and critical for real-time applications.

\begin{figure}[t]
\centering
\includegraphics[width=1\linewidth]{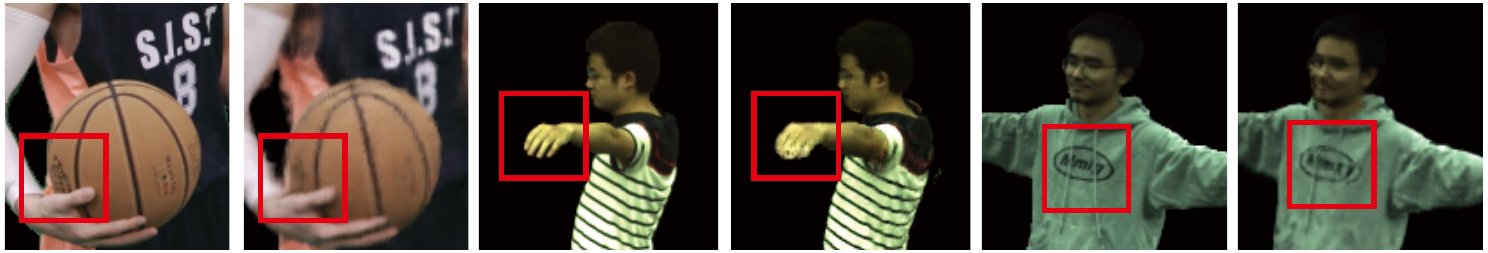}
\caption{\textbf{Failure cases.} Our method has difficulty in rendering very detailed content, such as the text and fingers.}
\label{fig:failure_cases}
\end{figure}

\PAR{Failure cases.}
Our method struggles to render very detailed content, such as the text and fingers. Figure~\ref{fig:failure_cases} presents some qualitative results.

\section{Societal impact}
The volumetric videos could be misused for recording and spreading some moments of real people without permission, which poses a threat to the privacy.
We strongly oppose such usage of our technique.

\section{Detailed results}

Tables \ref{tab:nhr} and \ref{tab:zju_mocap} present the per-scene comparison.
The results demonstrate that our proposed approach significantly outperforms baseline methods.

\begin{table*}[t]
	\centering
	\begin{tabular}{l|cccc}
		\toprule
        &   sport1 & sport2 & sport3 & basketball \\ \midrule
		Metric & \multicolumn{4}{c}{PSNR$\uparrow$  }\\ \midrule

	    Neural Volumes & 31.76 & 31.48 & 31.04 & 29.17 \\
	    C-NeRF & 31.81 & 32.12 & 31.99 &\textbf{29.35}  \\
	    D-NeRF & 30.12 & 30.18 & 29.66 & 27.02  \\
	    DyNeRF & 31.76 & 32.43 & 31.33 & 27.97 \\
	    Ours & \textbf{32.92} & \textbf{33.19} & \textbf{33.59} & 29.11 \\ \midrule

		Metric & \multicolumn{4}{c}{SSIM$\uparrow$ }\\\midrule

		Neural Volumes & 0.951 & 0.933 & 0.940 & 0.938 \\
	    C-NeRF & 0.954 & 0.950 & 0.950 & 0.942 \\
	    D-NeRF & 0.934 & 0.917 & 0.914 & 0.914  \\
	    DyNeRF & 0.954 & 0.945 & 0.944 & 0.929  \\
	    Ours & \textbf{0.959} & \textbf{0.954} &  \textbf{0.956} & \textbf{0.943} \\ \midrule

		Metric & \multicolumn{4}{c}{LPIPS $\downarrow$}\\ \midrule
        Neural Volumes & 0.106 & 0.143 & 0.131 & 0.141 \\
	    C-NeRF & 0.085 & 0.107 & 0.097 & 0.118 \\
	    D-NeRF & 0.111 & 0.169 & 0.155 & 0.163  \\
	    DyNeRF & 0.095 & 0.119 & 0.114 & 0.142 \\
	    Ours & \textbf{0.067} & \textbf{0.084} & \textbf{0.076} & \textbf{0.094} \\ \bottomrule
	
	\end{tabular}
	\caption{\textbf{Quantitative comparisons on the NHR dataset.}}

\label{tab:nhr}
\end{table*}

\begin{table*}[t]
	\centering
	\begin{tabular}{l|ccccccccc}
		\toprule
        &   313 & 315 & 377 & 386 & 387 & 390 & 392 & 393 & 394 \\ 
		\midrule
		Metric & \multicolumn{9}{c}{PSNR$\uparrow$  }\\ \midrule

	    Neural Volumes & 30.23 & 26.92 & 26.90 & 30.15 & 25.84 & 28.06 & 28.76 & 27.95 & 28.28 \\
	    C-NeRF & 31.84 & 29.03 & 28.92 & 30.75 & 27.52 & 29.81 & 30.82 & 29.62 & 29.81 \\
	    D-NeRF & 27.48 & 26.68 & 26.20 & 28.78 & 25.53 & 28.10 & 27.37 & 26.14 & 27.47  \\
	    DyNeRF & 31.50 & \textbf{30.29} & 28.92 & 30.88 & \textbf{27.90} & \textbf{30.14} & 30.09 & 29.28 & 29.88 \\
	    Ours & \textbf{32.15} & 29.94 & \textbf{29.40} & \textbf{31.05} & 27.89 & 30.10 & \textbf{31.06} & \textbf{29.78} & \textbf{30.15} \\ \midrule

		Metric & \multicolumn{9}{c}{SSIM$\uparrow$ }\\\midrule

		Neural Volumes & 0.949 & 0.947 & 0.921 & 0.949 & 0.910 & 0.924 & 0.939 & 0.936 & 0.932 \\
	    C-NeRF & 0.973 & 0.968 & 0.958 & 0.958 & 0.952 & 0.957 & 0.959 & 0.955 & 0.952 \\
	    D-NeRF & 0.927 & 0.939 & 0.920 & 0.937 & 0.918 & 0.933 & 0.911 & 0.897 & 0.915  \\
	    DyNeRF & 0.970 & 0.976 & 0.960 & \textbf{0.960} & \textbf{0.953} & \textbf{0.959} & 0.953 & 0.952 & 0.952 \\
	    Ours & \textbf{0.976} & \textbf{0.977} & \textbf{0.963} & \textbf{0.960} & \textbf{0.953} & \textbf{0.959} & \textbf{0.962} & \textbf{0.958} & \textbf{0.957} \\ \midrule

		Metric & \multicolumn{9}{c}{LPIPS $\downarrow$}\\ \midrule
        Neural Volumes & 0.103 & 0.114 & 0.147 & 0.122 & 0.169 & 0.149 & 0.127 & 0.128 & 0.124 \\
	    C-NeRF & 0.057 & 0.076 & 0.079 & \textbf{0.072} & \textbf{0.080} & 0.071 & 0.077 & 0.086 & 0.090 \\
	    D-NeRF & 0.122 & 0.106 & 0.150 & 0.132 & 0.140 & 0.119 & 0.166 & 0.162 & 0.150  \\
	    DyNeRF & 0.070 & 0.061 & 0.083 & 0.082 & 0.094 & 0.083 & 0.113 & 0.102 & 0.099 \\
	    Ours & \textbf{0.049} & \textbf{0.047} & \textbf{0.069} & \textbf{0.072} & 0.081 & \textbf{0.068} & \textbf{0.074} & \textbf{0.080} & \textbf{0.072} \\ \bottomrule
	\end{tabular}
	\caption{\textbf{Quantitative comparisons on the ZJU-MoCap dataset.}}

\label{tab:zju_mocap}
\end{table*}

\end{document}